%% file: 0.main.tex
\documentclass[sigconf, nonacm]{acmart}

\AtBeginDocument{%
  }

\usepackage{lipsum}
\usepackage{booktabs} 
\usepackage{url} 
\usepackage{caption}
\usepackage{pdfpages}
\usepackage{multirow}
\usepackage{kotex}
\usepackage{subcaption} 
\usepackage{listings}
\usepackage{xcolor}
\usepackage{subcaption}
\usepackage{adjustbox} 
\usepackage{subcaption}
\usepackage{enumitem}
\usepackage{dirtree}
\usepackage{mdframed}

\definecolor{codebg}{RGB}{245,245,244}
\definecolor{keyword}{RGB}{0,0,180}
\definecolor{comment}{RGB}{0,128,0}
\definecolor{string}{RGB}{163,21,21}

\lstdefinestyle{pythonstyle}{
  language=Python,
  keywordstyle=\color{blue},
  stringstyle=\color{red},
  commentstyle=\color{gray},
  numbers=left,
  numberstyle=\tiny,
  stepnumber=1,
  frame=single,
  breaklines=true,
  xleftmargin=1em,   
  xrightmargin=1em, 
  framexrightmargin=0pt 
}

\lstdefinestyle{bashstyle}{
  language=bash,
  backgroundcolor=\color{codebg},
  keywordstyle=\bfseries\color{keyword},
  commentstyle=\itshape\color{comment},
  stringstyle=\color{string},
  frame=single,
  framesep=6pt,
  frameround=tttt,
  numberstyle=\tiny\color{gray},
  stepnumber=1,
  showstringspaces=false,
  tabsize=2,
  breaklines=true,
  breakatwhitespace=true,
  captionpos=b,
  aboveskip=6pt, 
  belowskip=6pt
}

\begin{document}

\title{
  Alibaba International E-commerce Product Search Competition
  DILAB Team Technical Report
}

\author{Hyewon Lee}
\thanks{
Presented at the 34th ACM International Conference on Information and Knowledge Management (CIKM'25), November 10--14, 2025, Seoul, Republic of Korea. \\
\textcopyright\ 2025 DILAB. Licensed under \href{https://creativecommons.org/licenses/by/4.0/}{CC BY 4.0}.
}
\orcid{0009-0003-9573-2916}
\affiliation{%
  \institution{Chungnam National University}
  \city{Daejeon}
  \country{Republic of Korea}
}
\email{noweyh927@g.cnu.ac.kr}

\author{Junghyun Oh}
\orcid{0009-0004-5798-2831}
\affiliation{%
  \institution{Chungnam National University}
  \city{Daejeon}
  \country{Republic of Korea}
}
\email{ojh7839@o.cnu.ac.kr}

\author{Minkyung Song}
\orcid{0009-0004-4007-5418}
\affiliation{%
  \institution{Chungnam National University}
  \city{Daejeon}
  \country{Republic of Korea}
}
\email{kyung@o.cnu.ac.kr}

\author{Soyoung Park}
\orcid{0009-0000-7755-0041}
\affiliation{%
  \institution{Chungnam National University}
  \city{Daejeon}
  \country{Republic of Korea}
}
\email{sypark1452@o.cnu.ac.kr}

\author{Seunghoon Han}
\orcid{0009-0008-3393-4797}
\affiliation{%
  \institution{Chungnam National University}
  \city{Daejeon}
  \country{Republic of Korea}
}
\email{tmdgns129@g.cnu.ac.kr}

\renewcommand{\shortauthors}{Team DILAB}

\input{sections/0.abstract}

\ccsdesc[500]{Information systems~Retrieval effectiveness}
\ccsdesc[500]{Computing methodologies~Supervised learning by classification}
\ccsdesc[500]{Computing methodologies~Information extraction}

\keywords{Multilingual Information Retrieval, E-commerce Search, Query-Category (QC), Query–Item (QI), Large Language Model (LLM).}

\maketitle

\input{sections/1.intro}
\input{sections/2.method}
\input{sections/3.exp}

\input{sections/4.con}


\bibliographystyle{ACM-Reference-Format}
\bibliography{reference}

\end{document}

%% file: sections/0.abstract.tex
\begin{abstract}
    This study presents the multilingual e-commerce search system developed by the DILAB team, which achieved 5th place on the final leaderboard with a competitive overall score of 0.8819, demonstrating stable and high-performing results across evaluation metrics. To address challenges in multilingual query–item understanding, we designed a multi-stage pipeline integrating data refinement, lightweight preprocessing, and adaptive modeling. The data refinement stage enhanced dataset consistency and category coverage, while language tagging and noise filtering improved input quality. In the modeling phase, multiple architectures and fine-tuning strategies were explored, and hyperparameters optimized using curated validation sets to balance performance across query-category (QC) and query–item (QI) tasks. The proposed framework exhibited robustness and adaptability across languages and domains, highlighting the effectiveness of systematic data curation and iterative evaluation for multilingual search systems. The source code is available at \url{https://github.com/2noweyh/DILAB-Alibaba-Ecommerce-Search}.
\end{abstract}

%% file: sections/1.intro.tex
\section{Introduction}
\label{sec:intro}
Alibaba International Digital Commerce (AIDC)\footnote{\url{https://www.alibabagroup.com}} operates AliExpress, Lazada, Daraz, Trendyol, and Mirivia, serving users in over 100 countries and 20 languages. The search engine is a core component for product discovery, and optimizing multilingual search quality is critical for user satisfaction and competitiveness.  
We address the problem of \textbf{Multilingual E-commerce Product Search}, focusing on two key tasks: the \textit{Query-Category (QC) relevance task}, which determines whether a query matches a product category, and the \textit{Query-Item (QI) relevance task}, which evaluates whether a query corresponds to a specific product listing. Together, these tasks form the foundation of multilingual e-commerce search.  

\begin{table}[ht]
\centering
\large
\caption{Per-language sample counts in the QC and QI train and test datasets, illustrating the distribution of data across languages for multilingual relevance evaluation.}
\vspace{-0.1cm}
\setlength{\abovecaptionskip}{5pt}
\setlength{\tabcolsep}{5pt}
\renewcommand{\arraystretch}{1.2}
\resizebox{0.7\linewidth}{!}{
\begin{tabular}{c|c|cc|cc}
\toprule
\multirow{2}{*}{\textbf{Language}} & \multirow{2}{*}{\textbf{Code}} & \multicolumn{2}{c|}{\textbf{QC}} & \multicolumn{2}{c}{\textbf{QI}} \\
 &  & Train & Test & Train & Test \\
\midrule
English      & en & 50k & 10k & 50k & 10k \\
Spanish      & es & 50k & 10k & 50k & 10k \\
French       & fr & 50k & 10k & 50k & 10k \\
Japanese     & ja & 50k & 10k & 45k & 10k \\
Korean       & ko & 50k & 10k & 45k & 10k \\
Portuguese   & pt & 50k & 10k & 50k & 10k \\
Thai         & th & --  & --  & 50k & 10k \\
Arabic       & ar & --  & 10k & --  & 10k \\
German       & de & --  & 10k & --  & 10k \\
Italian      & it & --  & 10k & --  & 10k \\
Polish       & pl & --  & 10k & --  & 10k \\
Indonesian   & id & --  & --  & --  & 10k \\
Vietnamese   & vn & --  & --  & --  & 10k \\
\bottomrule
\end{tabular}
}
\label{tab:train_dev_test_lang_dist}
\end{table}

\begin{figure}[t]
\centering
\includegraphics[width=0.90\linewidth]{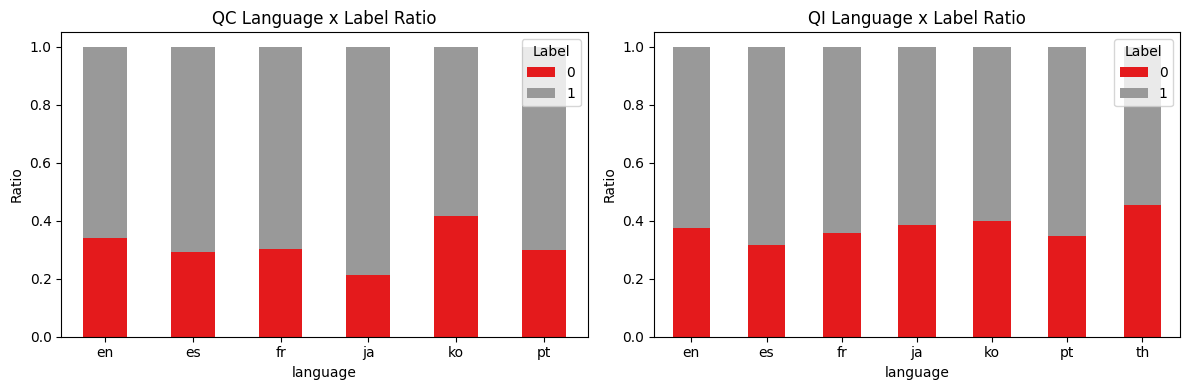}
\vspace{-0.4cm}
\caption{Label ratio distribution across languages in the QC and QI training datasets.}
\label{fig:lang_label_dist}
\end{figure}

Meanwhile, the datasets used in this competition — namely, the QC and QI datasets — were constructed from multilingual search logs and annotated by domain experts. Each dataset includes the query language, original query text, a corresponding element (either a category path or an item title), and a binary relevance label.
Table~\ref{tab:train_dev_test_lang_dist} summarizes key statistics of the QC and QI datasets, indicating that some languages appear only in the dev and test sets but not in the training data. Figure~\ref{fig:lang_label_dist} further illustrates a mild label imbalance and variations in language distribution across the datasets; however, neither issue was severe enough to require correction~\cite{ghosh2024class, johnson2019survey, van2022harm}.

This finding underscores the importance of multilingual processing and the ability to generalize to unseen languages. To address this challenge, large language models (LLMs) with strong cross-lingual transfer and generalization capabilities have proven particularly effective. However, the performance of LLMs depends not only on the quantity of available data but, more critically, on its quality. Prior studies have emphasized that massive amounts of high-quality data are essential for effective LLM training~\cite{zhang2024gpt}, and have further shown that systematic data filtering can yield better performance than using raw data indiscriminately~\cite{li2024superfiltering}. In line with these findings, our study highlights that data quality management and refinement are not optional but indispensable for successful model development.

To address this, our team applied lightweight preprocessing and modeling strategies, including Quality Refinement (QR) to filter out low-quality data and enhance overall dataset quality, and Language Tagging (LT) for explicit multilingual signals. Task-specific strategies, such as Hierarchical Category Tagging (HCT) in QC, and Semantic Item Tagging (SIT) and Description Generation (DG) in QI, were also deployed. In addition, we incorporated advanced modeling techniques such as In-context Learning (ICL) and Chain-of-Thought (CoT) reasoning to strengthen the model’s ability to generalize across languages and complex query structures. Together, these strategies are designed to enhance the model’s cross-lingual relevance prediction, consistent with prior studies showing the effectiveness of query augmentation for improving multilingual search performance~\cite{cakir2023modified, wang2025csrm}.


\vspace{-0.2cm}
\begin{figure}[ht]
\centering
\begin{minipage}{0.30\linewidth}
    \centering
    \includegraphics[width=1.0\linewidth]{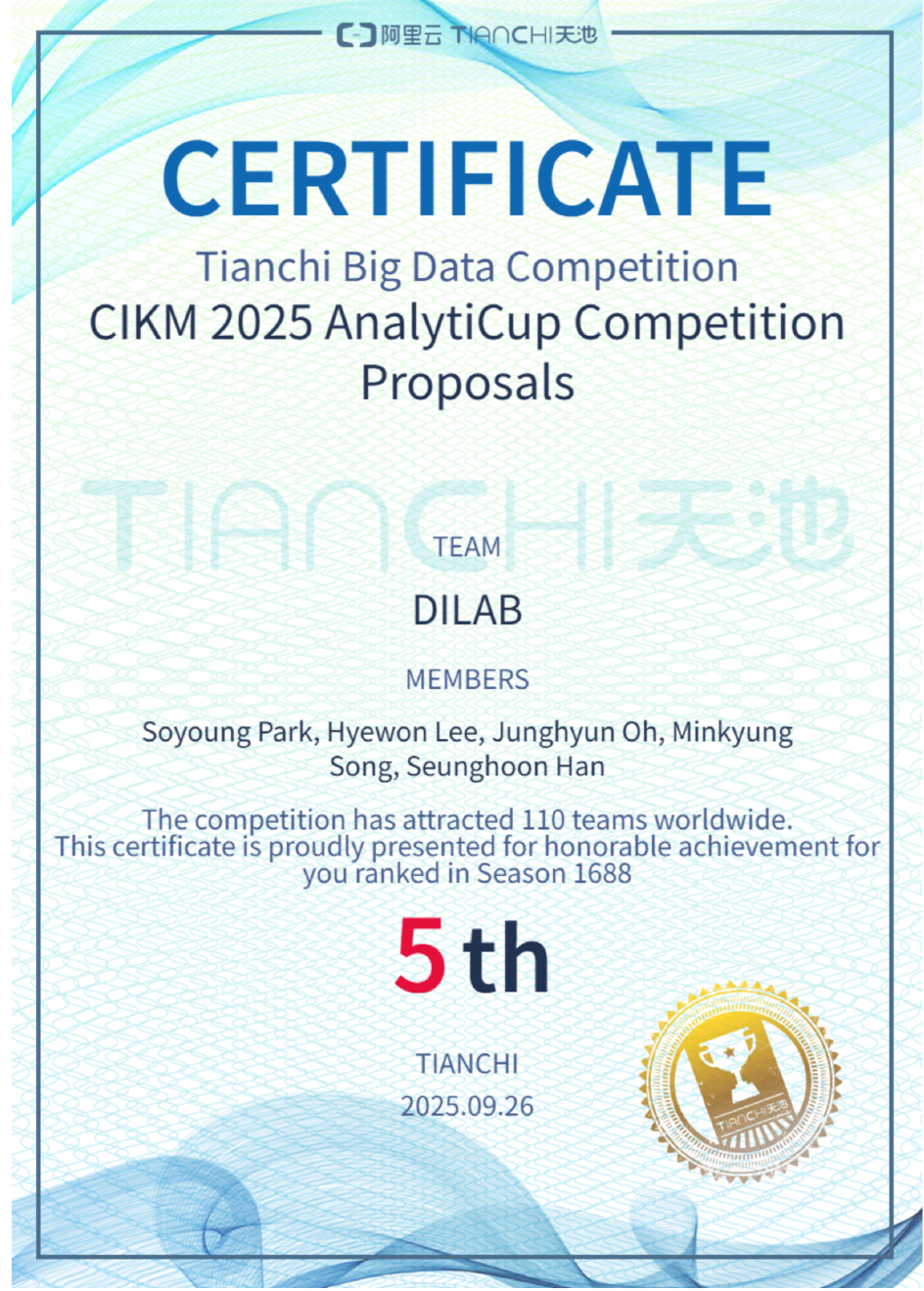}
    \label{fig:certificate}
\end{minipage}
\hfill
\begin{minipage}{0.655\linewidth}
    We focus on developing and applying LLMs that can handle multilingual data in e-commerce search. Using the QC and QI datasets, we leverage the multilingual capabilities of a pre-trained LLM with lightweight preprocessing strategies to improve relevance prediction across languages and tasks, ultimately enhancing search quality in global e-commerce environments. Our team advanced to the final round and achieved 5th place on the final leaderboard.
\end{minipage}
\end{figure}

\vspace{-0.3cm}
Our main contributions can be summarized as follows:
\begin{itemize}
    \item We propose an efficient multilingual search framework that combines lightweight preprocessing with LLMs to improve cross-lingual relevance prediction for both QC and QI tasks.
    
    \item We introduce task-specific enhancements, including HCT for QC and SIT with DG for QI, which enhance data quality and model interpretability.
    
    \item Our team’s rigorous approach to data tagging and modeling was recognized with a \textit{Special Award} at the Alibaba International E-commerce Product Search Competition, highlighting the originality and impact of our work.
\end{itemize}

The following sections describe our methodology and experiments in detail. Section~\ref{sec:method} presents the proposed approach based on the insights above, Section~\ref{sec:exp} outlines the experimental setup and results, and Section~\ref{sec:con} concludes with findings and future directions.

%% file: sections/2.method.tex
\section{Methodology}
\label{sec:method}

\begin{figure*}[t]
\centering
\includegraphics[width=0.8\linewidth]{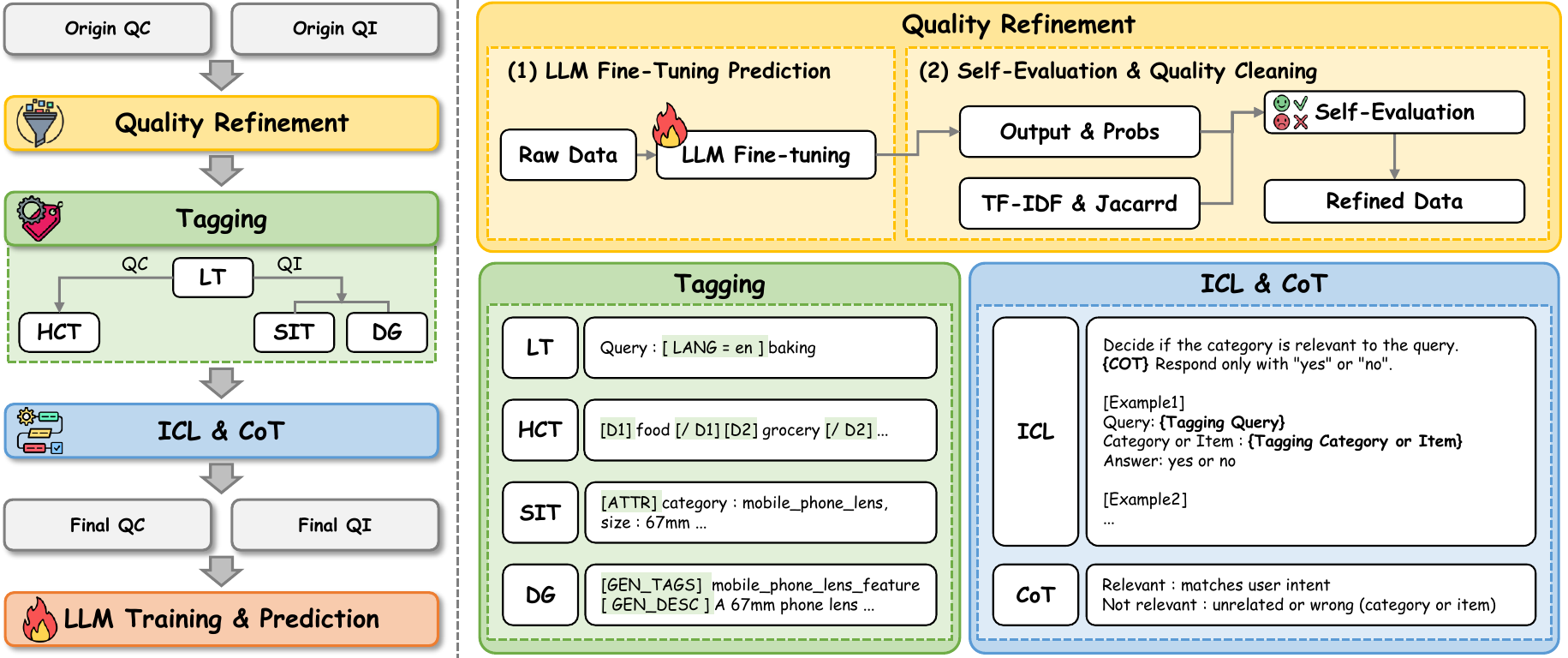}
\caption{Overview of the proposed DILAB framework for the QC and QI tasks.}
\label{fig:framework}
\end{figure*}

In this section, we present the overall methodology for the QC and QI tasks in multilingual e-commerce search. 
Our approach follows a unified pipeline. This pipeline, illustrated in Figure~\ref{fig:framework}, is composed of the following four key components:
\begin{enumerate}
    \item \textbf{Quality Refinement (QR):} filters out noisy or low-quality samples to improve dataset reliability.  
    \item \textbf{Tagging:} injects explicit linguistic and structural cues to enhance representational richness.  
    \item \textbf{In-Context Learning (ICL)} and \textbf{Chain-of-Thought (CoT):} strengthen multilingual reasoning and generalization.  
    \item \textbf{LLM Training and Prediction:} integrates all components into a unified multilingual relevance model.  
\end{enumerate}
This pipeline is specifically designed to improve both data quality and the generalization ability of large language models across diverse languages and product domains.

\subsection{Quality Refinement (QR)}  
To improve dataset quality, we designed a self-evaluation refinement that integrates model predictions with lexical similarity measures. The QR process operates in two steps: (1) LLM Fine-Tuning and Prediction; and (2) Self-Evaluation and Quality Cleaning. 

This refinement pipeline ensures that mislabeled or noisy data are filtered out before downstream training.
By embedding TF-IDF/Jaccard–based lexical similarity within a self-evaluation phase, the model not only learns from the data but also validates and improves data quality.
The proposed refinement strategy significantly enhances robustness and cross-lingual adaptability, making the refined datasets more reliable for multilingual e-commerce search tasks, and serving as a solid foundation for the subsequent tagging, ICL/CoT, and LLM training stages.

\subsubsection{LLM Fine-Tuning and Prediction} 
Raw QC and QI data are reformatted into lightweight instruction–input pairs and used to fine-tune the eCeLLM-M~\cite{peng2024ecellm} model with LoRA adaptation. 
In this step, we adopt the basic version of the prompt, which excludes \textit{Language Tagging}, \textit{Semantic Item Tagging}, and \textit{Description Generation}, unlike the enriched prompt format used in the final stage. The fine-tuned model then re-predicts the same training data, producing probabilities and binary relevance labels. 

\subsubsection{Self-Evaluation and Quality Cleaning}  
The probabilities and binary relevance labels from the previous step are combined with lexical similarity features—specifically TF-IDF cosine similarity and Jaccard overlap—between queries and paired categories/items. Language-specific thresholds are automatically tuned, and keyword-based guards mitigate false positive/negative bias. Samples with strong disagreement between labels, predictions, and similarity scores are flagged as suspect and removed. The final refined dataset is saved in JSONL format while preserving the original schema. 



\subsection{Tagging}  
\begin{figure}[ht]
  \centering
\begin{lstlisting}[style=pythonstyle]
# QC Prompt
Instruction:
You are given a user query 
and a product category.
Decide if the category is 
relevant to the query.
Relevant: matches user intent 
(correct category).
Not relevant: unrelated or 
wrong category.
Respond only with "yes" or "no".

Query: [LANG=en] baking
Category: [D1] food [/D1] 
[D2] grocery [/D2] 
[D3] flour [/D3] 
[D4] baking and cooking [/D4] 
[D5] decorations [/D5]
Answer: yes
Options: {In-Context Learning}
\end{lstlisting}
  \vspace{-0.2cm}
  \caption{Representative QC prompt structure.}
  \label{fig:qc_prompt}
\end{figure}

\begin{figure}[ht]
\centering
\begin{lstlisting}[style=pythonstyle]
# QI Prompt
Instruction:
You are given a user query 
and a product title.
Decide if the product is 
relevant to the query.
Relevant: matches user intent 
(category/type match, 
brand/specs may differ).
Not relevant: unrelated type 
or accessory instead of 
main item.
Respond only with "yes" or "no".

Query: [LANG=en] 
Apexel 60x telephoto lens
Product: 67mm Phone Filter 
Holder Mount Lens Filter Clip 
With Cold Shoe Adapter s
Universal For IPhone 
Photography Camera Accessories 
[ATTR] category: mobile_phone_lens, 
size: 67mm, qc_leaf: mobile_phone_lens
[GEN_TAGS] mobile_phone_lens_feature
[GEN_DESC] A 67mm phone lens 
filter holder mount for 
iPhone photography accessories.
Answer: no
Options: {In-Context Learning}
\end{lstlisting}
  \vspace{-0.2cm}
  \caption{Representative QI prompt structure.}
  \label{fig:qi_prompt}
\end{figure}

Prior studies in e-commerce have demonstrated that explicit tagging of hierarchical categories and item attributes improves fine-grained relevance modeling and retrieval performance~\cite{zhang2022semantic}.  
Motivated by these findings, we designed task-specific tagging schemes. We define four complementary tagging strategies as follows: 
\begin{itemize}
    \item \textbf{Language Tagging (LT):}   
    Both QC and QI datasets were annotated with language tags (e.g., Arabic queries marked as [LANG=ar]) to enhance multilingual recognition.  
    
    \item \textbf{Hierarchical Category Tagging (HCT):}  
    QC data are category-driven and hierarchically organized. We introduce structured tags such as [D1] apparel accessories [/D1] to make the depth explicit, enabling the model to distinguish detailed category levels.  
    A representative HCT prompt is shown in Figure~\ref{fig:qc_prompt}.  
    
    \item \textbf{Semantic Item Tagging (SIT):}  
    QI data require richer annotation.  
    We adopt a hybrid pipeline that combines rule-based attribute tagging (brand, color, size, material, style, etc.) with LLM-based attribute enrichment.  
    Detected attributes are standardized as [ATTR] tags.  
    
    \item  \textbf{Description Generation (DG):}  
    To further enrich the input representation, an instruction-tuned LLM (Meta-Llama-3-8B-Instruct~\cite{grattafiori2024llama} via vLLM) generates short descriptive sentences and binary relevance labels in a controlled format.  
    A representative SIT+DG prompt is shown in Figure~\ref{fig:qi_prompt}.
\end{itemize}

\subsection{In-Context Learning and Chain-of-Thought} 
According to previous studies in the e-commerce domain, integrating ICL and CoT has been shown to improve model reasoning and relevance prediction~\cite{sachdev2024automated}. 
Building on this finding, we applied ICL and CoT in our fine-tuning process. 
Specifically, ICL provided multilingual positive and negative examples to guide contextual understanding, while CoT offered explicit step-by-step reasoning cues that defined the relevance criteria, as shown in Figure~\ref{fig:framework}. 
This approach proved effective in enhancing the fine-tuning process. 
Figures~\ref{fig:qc_prompt} and~\ref{fig:qi_prompt} illustrate CoT guidelines for each task, and the overall context of both CoT and ICL can also be observed in Figure~\ref{fig:framework}.

\subsection{LLM Training and Prediction}
We experimented with several backbone models used in e-commerce research, including Qwen2.5-14B-Instruct~\cite{yang2025qwen3}, Meta-LLaMA-3-8B-Instruct~\cite{grattafiori2024llama}, and eCeLLM~\cite{peng2024ecellm}. 
Among them, Qwen2.5-14B-Instruct achieved the best performance across the QC and QI tasks.

We then performed task-specific fine-tuning on this instruction-tuned backbone. This process was unified across the QC and QI tasks for consistent optimization, and parameter-efficient adaptation was performed via LoRA~\cite{hu2022lora}. To ensure reproducibility, the fine-tuned model was applied to the development and test inputs using the same setup as in training. This process yields the final, optimized model used for all subsequent evaluations.


%% file: sections/3.exp.tex
\section{Experiments}
\label{sec:exp}
To evaluate the effectiveness of the proposed method, we conducted experiments on the provided test dataset.
This section presents the experimental settings, implementation, and results, including data preprocessing, model training, prediction, and ablation analysis.

\subsection{Requirements}
All experiments were conducted in a Conda-based Python 3.11 environment. 
Table~\ref{tab:env} summarizes the hardware and software setup used throughout the experiments.

\begin{table}[ht]
\centering
\caption{Experimental environment setup}
\vspace{-0.2cm}
\setlength{\abovecaptionskip}{5pt}
\setlength{\tabcolsep}{5pt}
\renewcommand{\arraystretch}{1.1}
\resizebox{0.85\columnwidth}{!}{
\begin{tabular}{p{2.5cm}|p{4.5cm}}
\toprule
\textbf{Component} & \textbf{Specification} \\
\midrule
OS & Linux \\
Python & 3.11 \\
PyTorch & 2.7.1 \\
Transformers & 4.44.0 \\
GPU & NVIDIA H100 PCIe (80GB VRAM) \\ 
CUDA / Driver & CUDA 12.4 / Driver 550.54.14 \\ 
\bottomrule
\end{tabular}
}
\label{tab:env}
\end{table}

\subsection{Implementation}
This subsection provides an overview of the implementation process of our proposed method. We describe the overall code structure, data preprocessing steps, model training procedure, and prediction pipeline. Each component is designed to ensure reproducibility and efficiency throughout the entire workflow.

\subsubsection{Code Structure}
\begin{figure}[ht]
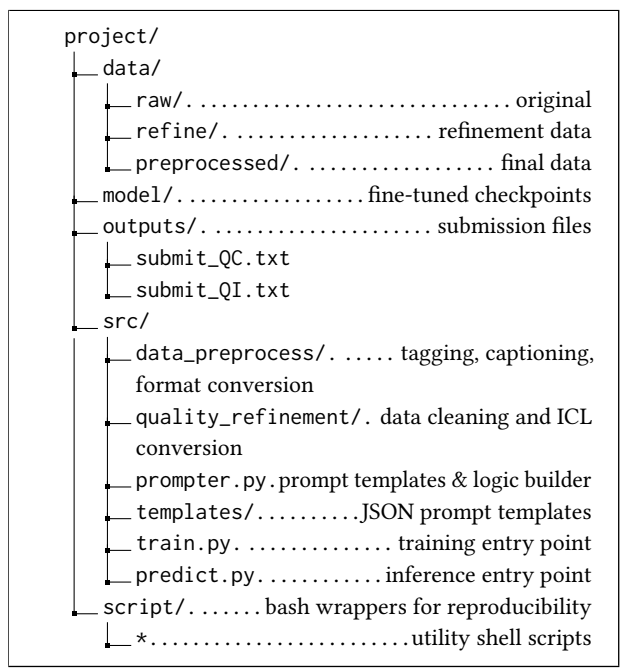

\centering
\begin{mdframed}[linewidth=0.5pt, roundcorner=6pt, innermargin=6pt, outermargin=4pt]
\dirtree{%
.1 project/.
.2 data/.
.3 raw/.\DTcomment{original}.
.3 refine/.\DTcomment{refinement data}.
.3 preprocessed/.\DTcomment{final data}.
.2 model/.\DTcomment{fine-tuned checkpoints}.
.2 outputs/.\DTcomment{submission files}.
.3 submit\_QC.txt.
.3 submit\_QI.txt.
.2 src/.
.3 data\_preprocess/.\DTcomment{tagging, captioning, format conversion}.
.3 quality\_refinement/.\DTcomment{data cleaning and ICL conversion}.
.3 prompter.py.\DTcomment{prompt templates \& logic builder}.
.3 templates/.\DTcomment{JSON prompt templates}.
.3 train.py.\DTcomment{training entry point}.
.3 predict.py.\DTcomment{inference entry point}.
.2 script/.\DTcomment{bash wrappers for reproducibility}.
.3 *.\DTcomment{utility shell scripts}.
}
\end{mdframed}
\caption{Project directory tree with descriptions.}
\label{fig:dir_tree}
\end{figure}

The directory is organized into modular components to ensure clarity and reproducibility, as shown in Figure~\ref{fig:dir_tree}.

\begin{itemize}
  \item \textbf{data/}: raw (original), refine (refinement data), preprocessed (final data).
  \item \textbf{model/}: fine-tuned checkpoints.
  \item \textbf{outputs/}: submission files (submit\_QC.txt, submit\_QI.txt).
  \item \textbf{src/data\_preprocess/}: tagging, captioning, format conversion.
  \item \textbf{src/quality\_refinement/}: data cleaning and ICL conversion.
  \item \textbf{src/prompter.py}, \textbf{src/templates/}: prompt templates \& logic.
  \item \textbf{src/train.py}, \textbf{src/predict.py}: training and inference entry points.
  \item \textbf{script/\*}: bash wrappers for reproducibility.
\end{itemize}

\subsubsection{Data Preprocessing} 
As shown in Listing~\ref{lst:qr-preprocess}, the QC and QI datasets are processed with an integrated script that converts raw data (\texttt{data/raw/}) into training and test sets through two stages: 
\textit{refinement}, which improves data quality, and 
\textit{preprocessing}, which applies tagging and format conversion.

\begin{center}
\begin{minipage}{0.95\columnwidth}
\begin{lstlisting}[style=bashstyle, caption={QR and Preprocessing commands.}, label={lst:qr-preprocess}]
# Quality Refinement (QR)
bash script/quality_refinement.sh 

# Preprocessing refined data
bash script/data_preprocess.sh         
\end{lstlisting}
\end{minipage}
\end{center}

\subsubsection{Training} 
As illustrated in Listing~\ref{lst:train-qc-qi}, model training is launched with the following arguments: 
\begin{itemize}
    \item \texttt{<num\_epochs>}: number of training epochs (e.g., 2).  
    \item \texttt{<model>}: backbone LLM (e.g., Qwen2.5-14B, Meta-Llama-3-8B, eCeLLM-M).  
    \item \texttt{<task>}: target task (\texttt{QC} or \texttt{QI}).  
\end{itemize}
The fine-tuned model checkpoints are saved in \texttt{./model/} directory.

\begin{center}
\begin{minipage}{0.95\columnwidth}
\begin{lstlisting}[style=bashstyle, caption={Training commands for QC and QI.},
                   label={lst:train-qc-qi},]
# Train QC and QI
# bash script/train.sh <num epochs> <model> <task>
bash script/train.sh 2 Qwen2.5-14B QC  
bash script/train.sh 2 Qwen2.5-14B QI 
\end{lstlisting}
\end{minipage}
\end{center}

\subsubsection{Prediction} 
As depicted in Listing~\ref{lst:predict-qc-qi}, inference is launched with the following arguments:  
\begin{itemize}
    \item \texttt{<model>}: backbone LLM (must match the fine-tuned model).  
    \item \texttt{<task>}: target task (\texttt{QC} or \texttt{QI}).  
\end{itemize}
This process produces two final submission files: \texttt{submit\_QC.txt} and \texttt{submit\_QI.txt}, which are located in the \texttt{./outputs/} folder.

\begin{center}
\begin{minipage}{0.95\columnwidth}
\begin{lstlisting}[style=bashstyle, caption={Prediction commands for QC and QI.},
                   label={lst:predict-qc-qi},]
# Predict QC and QI
# bash script/predict.sh <model> <task>
bash script/predict.sh Qwen2.5-14B QC
bash script/predict.sh Qwen2.5-14B QI
\end{lstlisting}
\end{minipage}
\end{center}

\subsection{Results}
\begin{table}[ht]
\centering
\caption{Comparison of overall F1 scores on the QC and QI datasets among LLM-based models.}
\vspace{-0.2cm}
\resizebox{0.85\columnwidth}{!}{
\begin{tabular}{c|c|c|c}
\toprule
\textbf{Method} & \textbf{Score\_QC} & \textbf{Score\_QI} & \textbf{Score} \\
\midrule
XLM-R & 0.8213 & 0.7936 & 0.8075 \\
Qwen2.5-1.5B & 0.8315 & 0.8137 & 0.8226 \\
Qwen2.5-7B & 0.8607 & 0.8487 & 0.8544  \\
Qwen2.5-14B & 0.8684 & 0.8667 & 0.8676 \\
\midrule
\textbf{Ours} & \textbf{0.8861} & \textbf{0.8778} & \textbf{0.8819} \\
\bottomrule
\end{tabular}
}
\label{tab:main_results}
\end{table}

We applied task-specific strategies that led to the performance improvements summarized in Table~\ref{tab:main_results}. Specifically, on the test set, our approach achieved gains of 0.0177 on QC and 0.0111 on QI, resulting in an overall improvement of 0.0143 compared to the previous best-performing model, Qwen2.5-14B. Furthermore, compared to the baseline XLM-R model provided in the competition baseline code, our strategy yielded improvements of 0.0648 on QC, 0.0842 on QI, and an overall gain of 0.0744. These results demonstrate the effectiveness of our proposed approaches.

\subsection{Ablation Study}

\begin{table}[ht]
\centering
\caption{Ablation studies on QC and QI datasets.}
\renewcommand{\arraystretch}{0.9}
\begin{minipage}[t]{0.48\linewidth}
\centering
\subcaption{QC dataset}
\resizebox{\linewidth}{!}{
\begin{tabular}{ccc|c}
\toprule
\textbf{QR} & \textbf{LT} & \textbf{HCT} & \textbf{Score} \\ 
\midrule
 & \checkmark &   & 0.8718 \\
 &  & \checkmark  & 0.8798 \\
 & \checkmark & \checkmark  & 0.8836 \\
\midrule
\checkmark & \checkmark & \checkmark & \textbf{0.8861} \\
\bottomrule
\end{tabular}
}
\label{tab:ablation_qc}
\end{minipage}
\hfill
\begin{minipage}[t]{0.48\linewidth}
\centering
\subcaption{QI dataset}
\resizebox{\linewidth}{!}{
\begin{tabular}{cccc|c}
\toprule
\textbf{QR} & \textbf{LT} & \textbf{SIT} & \textbf{DG} & \textbf{Score} \\ 
\midrule
 & \checkmark &  &  & 0.8701 \\
 &  &  \checkmark &  & 0.8741 \\
 &  &  & \checkmark & 0.8736 \\
 & \checkmark & \checkmark & \checkmark & 0.8751 \\
 \midrule
\checkmark & \checkmark & \checkmark & \checkmark & \textbf{0.8778} \\
\bottomrule
\end{tabular}
}
\label{tab:ablation_qi}
\end{minipage}
\end{table}

We evaluate the effectiveness of the proposed strategies as presented in Tables $\ref{tab:ablation_qc}$ and $\ref{tab:ablation_qi}$. It is observed that QR, applied consistently across QC and QI, yields uniform performance improvements, substantiating the critical importance of foundational data quality. For the QC task, HCT has a greater individual impact on performance than LT. Furthermore, the combination of QR, LT, and HCT achieves the highest performance of 0.8861, demonstrating a strong additive effect among the strategies in the QC domain. Similarly, for the QI task, the individual use of SIT shows the strongest result compared to LT and DG. Crucially, applying all four strategies (QR, LT, SIT, and DG) achieves the highest overall performance of 0.8778, demonstrating that maximum performance in the QI task is also realized through the synergistic application of the full set of preprocessing and modeling enhancements.


\subsection{Parameter Limitation}
As outlined above, we adopt task-specific strategies that lead to improved predictive performance, while employing a model comprising 14.98 billion parameters, thereby remaining compliant with the competition’s regulation limit of 15 billion parameters.

%% file: sections/4.con.tex
\section{Conclusion and Future Work}
\label{sec:con}
Our research successfully demonstrated the feasibility of using large language models (LLMs) for the Multilingual E-commerce Product Search problem, securing a top-tier rank in the preliminary round. This was achieved through a thorough understanding of the task and carefully designed strategies for data preprocessing, training, and inference, which ensured both computational and time efficiency. Importantly, these approaches enabled the model to remain robust when evaluated on test datasets containing previously unseen languages, confirming its effectiveness in challenging multilingual settings. Ultimately, we fulfilled our original goal of showing that LLM-based methods can reliably advance multilingual product search by achieving accurate and stable performance in noisy, real-world e-commerce environments.  

Overall, this work confirms that LLM-based approaches provide practical and reliable solutions for multilingual product search, bridging the gap between large-scale model capability and domain-specific adaptation. While our current system already achieves robust performance across diverse languages and product domains, future work will focus on further streamlining the fine-tuning process to reduce computational cost and exploring lightweight adaptation techniques to enable scalable real-world deployment.